\documentclass[11pt]{article}
\usepackage{acl2016}
\usepackage{times}
\usepackage{latexsym}
\usepackage{multirow}
\usepackage{fixltx2e}
\usepackage{graphicx}
\usepackage{caption}
\usepackage{subcaption}
\usepackage{amsmath}
\usepackage{amsthm}
\usepackage{amssymb}
\usepackage{xspace}
\usepackage{url}
\usepackage{wrapfig}
\usepackage[utf8]{inputenc}
\usepackage[normalem]{ulem}

\aclfinalcopy % Uncomment this line for the final submission
\title{Neural Machine Transliteration: Preliminary Results}

\author{Amir H. Jadidinejad\\
	    Bayan Inc.\\
	    {\tt jadidinejad@bayan.co.ir}}

\date{}

\begin{document}

\maketitle

\begin{abstract}
Machine transliteration is the process of automatically transforming the script of a word from a source language to a target language, while preserving pronunciation. Sequence to sequence learning has recently emerged as a new paradigm in supervised learning.
In this paper a character-based encoder-decoder model has been proposed that consists of two Recurrent Neural Networks. The encoder is a Bidirectional recurrent neural network that encodes a sequence of symbols into a fixed-length vector representation, and the decoder generates the target sequence using an attention-based recurrent neural network. The encoder, the decoder and the attention mechanism are jointly trained to maximize the conditional probability of a target sequence given a source sequence. Our experiments on different datasets show that the proposed encoder-decoder model is able to achieve significantly higher transliteration quality over traditional statistical models. 
\end{abstract}

\section{Introduction}
\label{sec:intro}
Machine Transliteration is defined as phonetic transformation of names across languages~\cite{news15white,machine_transliteration_survey11}. Transliteration of named entities is the essential part of many multilingual applications, such as machine translation~\cite{smt10} and cross-language information retrieval~\cite{CLIR_meta10}.

Recent studies pay a great attention to the task of Neural Machine Translation~\cite{nmt_properties14,seq2seq14}. In neural machine translation, a single neural network is responsible for reading a source sentence and generates its translation. 
From a probabilistic perspective, translation is equivalent to finding a target
sentence $\mathbf{y}$ that maximizes the conditional probability of $\mathbf{y}$ given a
source sentence $\mathbf{x}$, i.e., $\arg\max_{\mathbf{y}}{p(\mathbf{y} \mid \mathbf{x})}$. The whole neural network is \emph{jointly} trained to maximize the conditional probability of a correct translation given a source sentence, using the bilingual corpus. 

Transforming a name from spelling to phonetic and then use the constructed phonetic to generate the spelling on the target language is a very complex task~\cite{transliteration_survey06,nn_transliteration15}. Based on successful studies on Neural Machine Translation~\cite{nmt_properties14,seq2seq14,nlp15}, in this paper, we proposed a character-based encoder--decoder model which learn to transliterate end-to-end. In the opposite side of classical models which contains different components, the proposed model is trained end-to-end, so it able to apply to any language pairs without tuning for a spacific one.

\section{Proposed Model}
\label{sec:model}
Here, we describe briefly the underlying framework, called {\it RNN
Encoder--Decoder}, proposed by \cite{learning_phrase_rnn14} and \cite{seq2seq14} upon
which we build a machine transliteration model that learns to transliterate end-to-end.

The enoder is a character-based recurrent neural network that learns a highly nonlinear mapping from a spelling to the phonetic of the input sequence. This network reads the source name $x=(x_1, \dots, x_{T})$ and encodes it
into a sequence of hidden states $h=(h_1, \cdots, h_{T})$:
\begin{equation}\label{eq:generic_encoder}
    h_t = f\left( x_{t}, h_{t-1} \right)
\end{equation}
Each hidden state $h_i$ is a bidirectional recurrent representation with forward and backward sequence information around the $i$th character. The representation of a forward sequence and a backward sequence of the input character sequence is estimated and concatenated to form a context set $C=\{h_1, h_2,..., h_T\}$~\cite{mtl_mt15,char_decoder16}.
Then, the decoder, another recurrent neural network, computes the conditional distribution over all possible transliteration based on this context set and generates the corresponding
transliteration $y=(y_1, \cdots, y_{T'})$ based on the encoded sequence of hidden
states $h$.

The whole model is jointly trained to maximize the conditional log-probability of
the correct transliteration given a source sequence with respect to the parameters
$\theta$ of the model:
\begin{equation}\label{eq:conditional_output}
    \mathbf{\theta}^* = \arg\max_{\theta} \sum_{n=1}^N \sum_{t=1}^{T_n} \log p(y_t^n \mid y_{<t}^n, x^n),
\end{equation}
where $(x^n, y^n)$ is the $n$-th training pair of character sequences, and $T_n$ is the
length of the $n$-th target sequence ($y^n$). For each conditional term in Equation~\ref{eq:conditional_output}, the decoder updates its hidden state by:
\begin{equation}\label{eq:generic_encoder}
    h_{t'} = f\left( y_{t'-1}, h_{t'-1}, c_{t'} \right)
\end{equation}
where $c_{t'}$ is a context vector computed by a soft attention mechanism:
\begin{equation}\label{eq:soft_align}
    c_{t'} = f_a\left( y_{t'-1}, h_{t'-1}, C \right)
\end{equation}
The soft attention mechanism $f_a$ weights each vector in the context set $C$ according to its relevance given what has been transliterated.

Finally, the hidden state $h_{t'}$ , together with the previous target symbol $y_{t'-1}$ and the context vector $c_{t'}$ , is fed into a feedforward neural network to result in the conditional distribution described in Equation~\ref{eq:conditional_output}. The whole model, consisting of the encoder, decoder and soft attention mechanism, is trained end-to-end to minimize the negative log-likelihood using stochastic gradient descent.

\begin{table}
\small
\centering
\begin{tabular}{|l||l|l|l|l|l|}
\hline 
\bf \multirow{2}{*}{TaskID} & \bf \multirow{2}{*}{Source} & \bf \multirow{2}{*}{Target} & \multicolumn{3}{ |c| }{\bf Data Size} \\ \cline{4-6}
  &  &  & \bf Train & \bf Dev & \bf Test \\ \hline
En-Ch & English & Chinese & 37K & 2.8K  & 1.008K \\
Ch-En & Chinese & English & 28K & 2.7K  & 1.019K \\
En-Th & English & Thai    & 27K & 2.0K  & 1.236K \\
Th-En & Thai    & English & 25K & 2.0K  & 1.236K \\
En-Hi & English & Hindi   & 12K & 1.0K  & 1.000K \\
En-Ta & English & Tamil   & 10K & 1.0K  & 1.000K \\
En-Ka & English & Kannada & 10K & 1.0K  & 1.000K \\
En-Ba & English & Bangla  & 13K & 1.0K  & 1.000K \\
En-He & English & Hebrew  & 9.5K & 1.0K & 1.100K \\
En-Pe & English & Persian & 10K & 2.0K  & 1.042K \\
\hline
\end{tabular}
\caption{\label{tbl:datasets} Datasets provided by NEWS 2015~\cite{news15report}.}
\end{table}

\section{Experiments}
\label{sec:experiments}

\begin{table*}
\small
\centering
\begin{tabular}{|l||l|l|l|l|l|l|l|l|}
\hline 
\bf \multirow{2}{*}{TaskID} & \multicolumn{4}{ |c| }{\bf Baseline} & \multicolumn{4}{ |c| }{\bf Neural Machine Transliteration} \\ \cline{2-9}
  & \bf ACC & \bf F-Score & \bf MRR & \bf MAP & \bf ACC & \bf F-Score & \bf MRR & \bf MAP \\ \hline
En-Ch & 0.1935 & 0.5851 & 0.1935 & 0.1830 & 0.2659 & 0.6227 & 0.3185 & 0.2549 \\
Ch-En & 0.0981 & 0.6459 & 0.0981 & 0.0953 & 0.0834 & 0.6564 & 0.1425 & 0.0830 \\
En-Th & 0.0680 & 0.7070 & 0.0680 & 0.0680 & 0.1456 & 0.7514 & 0.2181 & 0.1456 \\
Th-En & 0.0914 & 0.7397 & 0.0914 & 0.0914 & 0.1286 & 0.7624 & 0.1966 & 0.1286 \\
En-Hi & 0.2700 & 0.7992 & 0.2700 & 0.2624 & 0.3480 & 0.8349 & 0.4745 & 0.3434 \\
En-Ta & 0.2580 & 0.8117 & 0.2580 & 0.2573 & 0.3240 & 0.8369 & 0.4461 & 0.3235 \\
En-Ka & 0.1960 & 0.7833 & 0.1960 & 0.1955 & 0.2860 & 0.8224 & 0.4019 & 0.2856 \\
En-Ba & 0.2870 & 0.8360 & 0.2870 & 0.2837 & 0.3460 & 0.8600 & 0.4737 & 0.3438 \\
En-He & 0.1091 & 0.7715 & 0.1091 & 0.1077 & 0.1591 & 0.7976 & 0.2377 & 0.1582 \\
En-Pe & 0.4818 & 0.9060 & 0.4818 & 0.4482 & 0.5816 & 0.9267 & 0.7116 & 0.5673 \\
\hline
\end{tabular}
\caption{\label{tbl:results} The effectiveness of neural machine transliteration is compared with the robust baseline~\cite{moses07} provided by NEWS 2016 shared task on transliteration of named entities.}
\end{table*}

We conducted a set of experiments to show the effectiveness of RNN Encoder--Decoder model~\cite{learning_phrase_rnn14,seq2seq14} in the task of machine transliteration using standard benchmark datasets provided by NEWS 2015-16 shared task~\cite{news15report}. Table~\ref{tbl:datasets} shows different datasets in our experiments. Each dataset covers different levels of difficulty and training set size. The proposed model has been applied on each dataset without tuning the algorithm for each specific language pairs. Also, we don't apply any preprocessing on the source or target language in order to evaluate the effectiveness of the proposed model in a fair situation. `TaskID' is a unique identifier in the following experiments.

We leveraged a character-based encoder--decoder model~\cite{alt_char_rnn15,char_decoder16} with soft attention mechanism~\cite{learning_phrase_rnn14}. In this model, input sequences in both source and target languages have been represented as characters. Using characters instead of words leads to longer sequences, so Gated Recurrent Units~\cite{nmt_properties14} have been used for the encoder network to model long term dependencies. The encoder has 128 hidden units for each direction (forward and backward), and the decoder has 128 hidden units with soft attention mechanism~\cite{learning_phrase_rnn14}. We train the model using stochastic gradient descent with Adam~\cite{adam14}. Each update is computed using a minibatch of 128 sequence pairs. The norm of the gradient is clipped with a threshold 1~\cite{deep_rnn13}. Also, beamsearch has been used to approximately find the most likely transliteration given a source sequence~\cite{smt10}.

Table~\ref{tbl:results} shows the effectiveness of the proposed model on different datasets using standard measures~\cite{news15report}. The proposed neural machine transliteration model has been compared to the baseline method provided by NEWS 2016 organizers~\cite{news15report}. Baseline results are based on a machine translation implementation at the character level using MOSES~\cite{moses07}. Experimental results shows that the proposed model is significantly better than the robust baseline using different metrics. 

\begin{figure*}
    \centering
    \begin{subfigure}[b]{0.3\textwidth}
        \includegraphics[width=\textwidth]{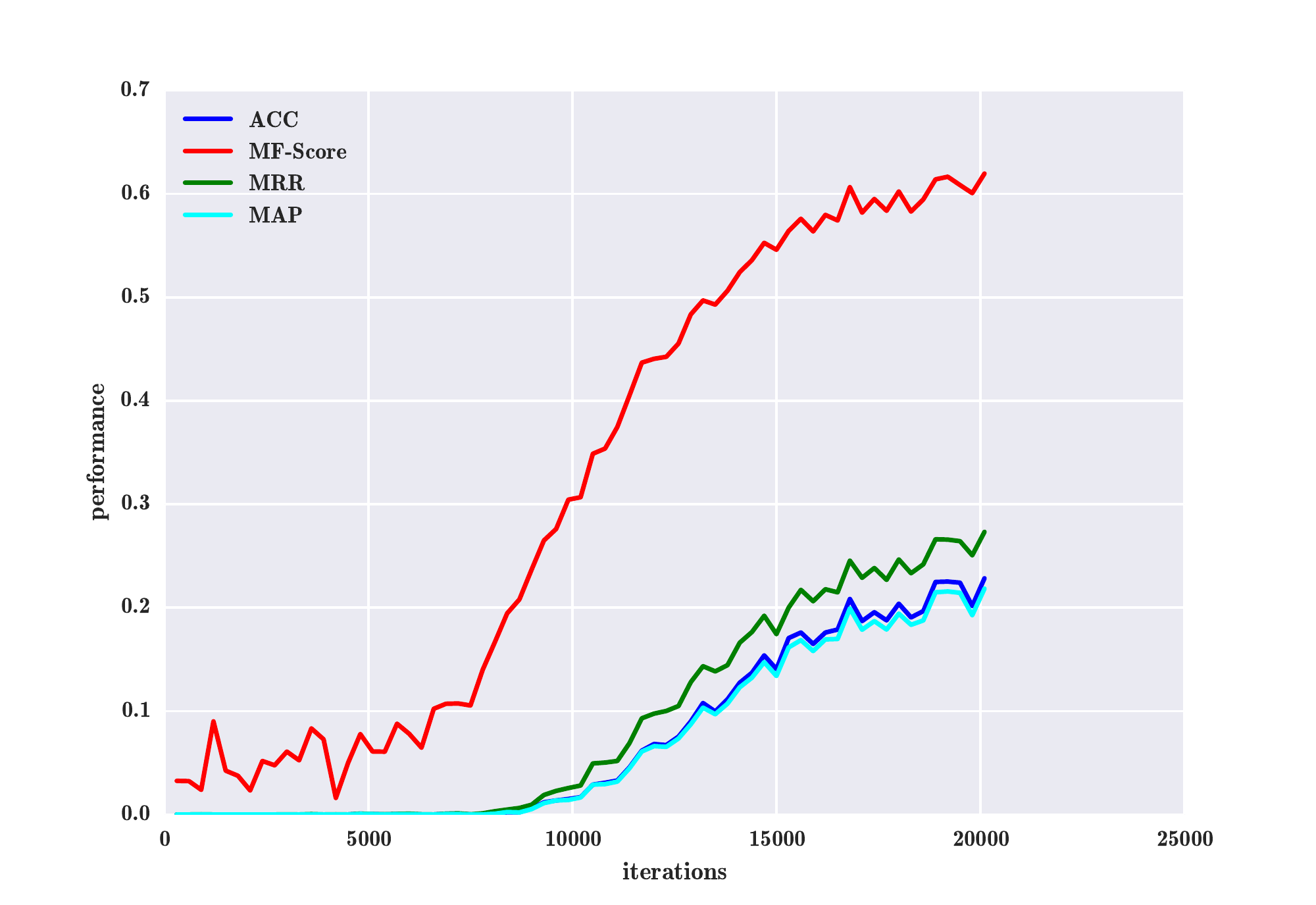}
        \caption{En-Ch}
        \label{fig:en-ch}
    \end{subfigure}
    \begin{subfigure}[b]{0.3\textwidth}
        \includegraphics[width=\textwidth]{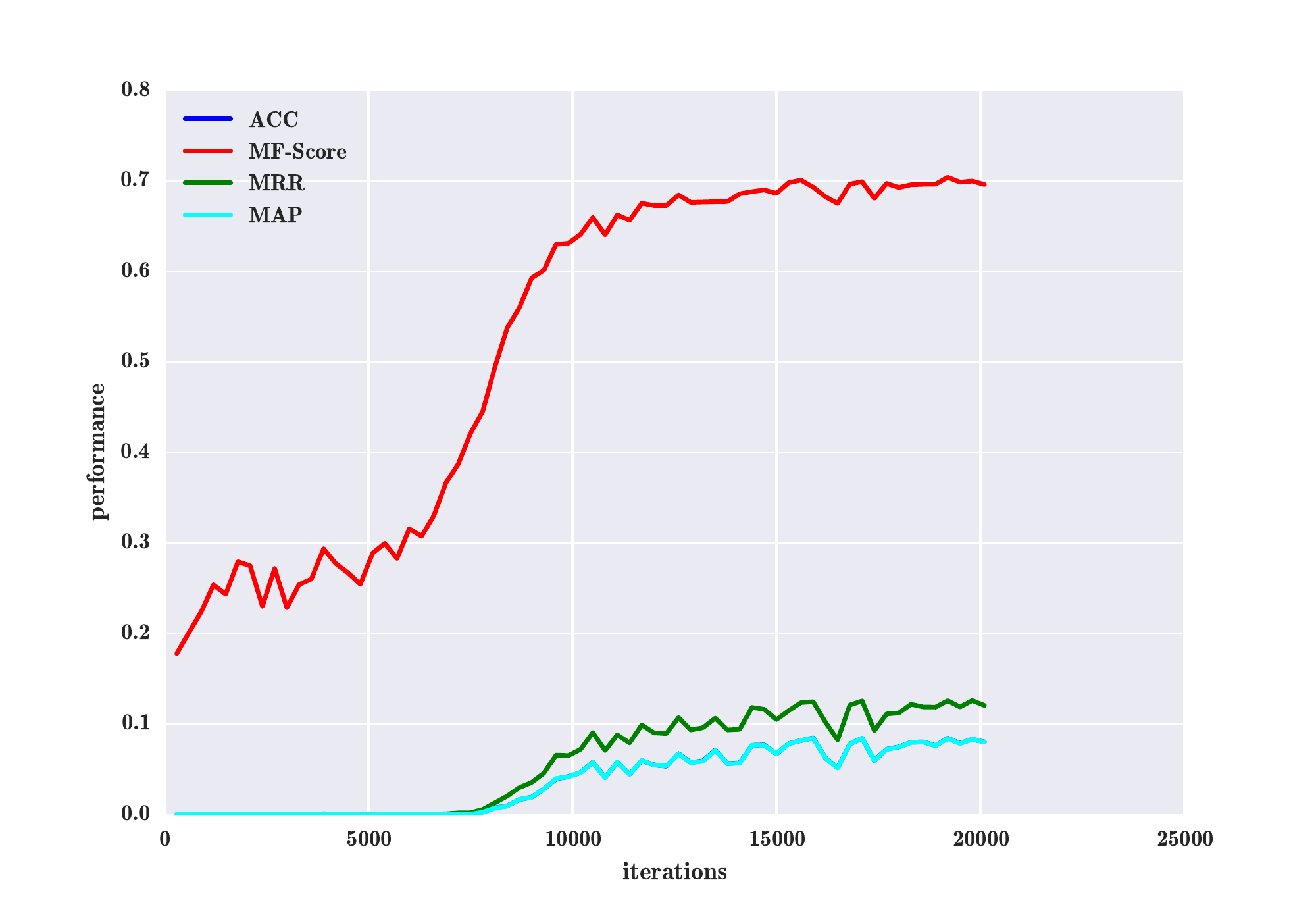}
        \caption{Ch-En}
        \label{fig:ch-en}
    \end{subfigure}
    \begin{subfigure}[b]{0.3\textwidth}
        \includegraphics[width=\textwidth]{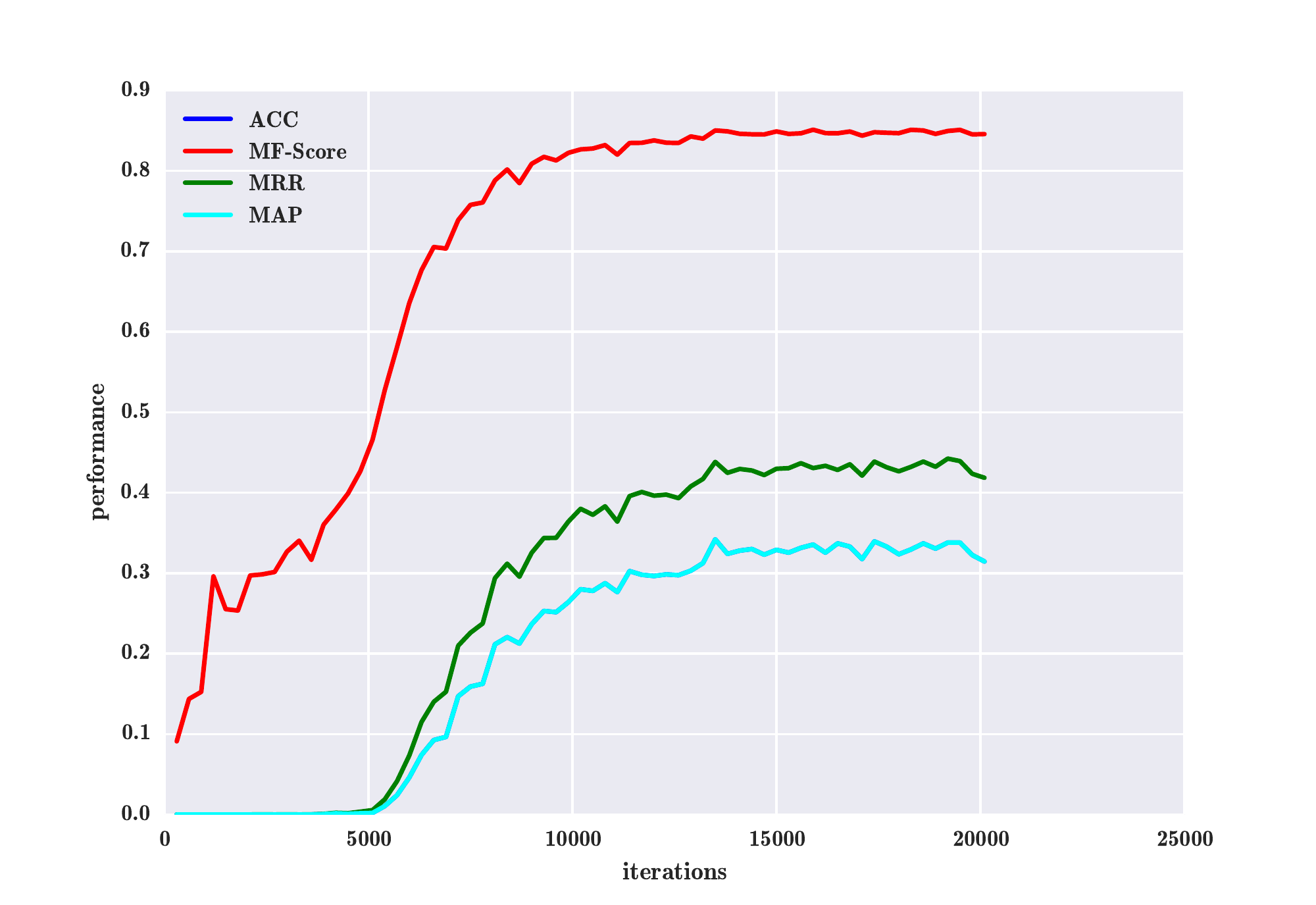}
        \caption{En-Th}
        \label{fig:en-th}
    \end{subfigure}
    \begin{subfigure}[b]{0.3\textwidth}
        \includegraphics[width=\textwidth]{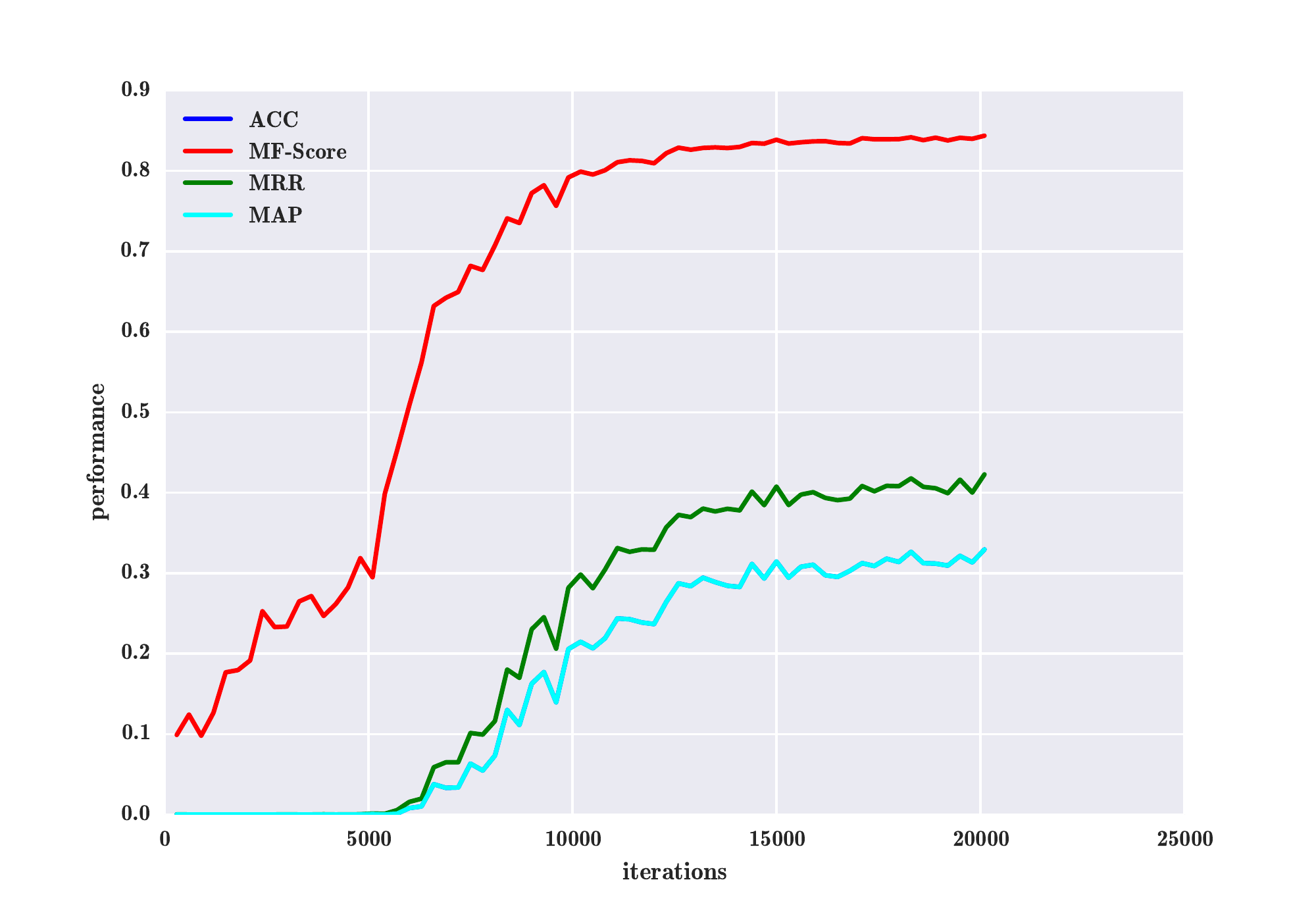}
        \caption{Th-En}
        \label{fig:th-en}
    \end{subfigure}
    \begin{subfigure}[b]{0.3\textwidth}
        \includegraphics[width=\textwidth]{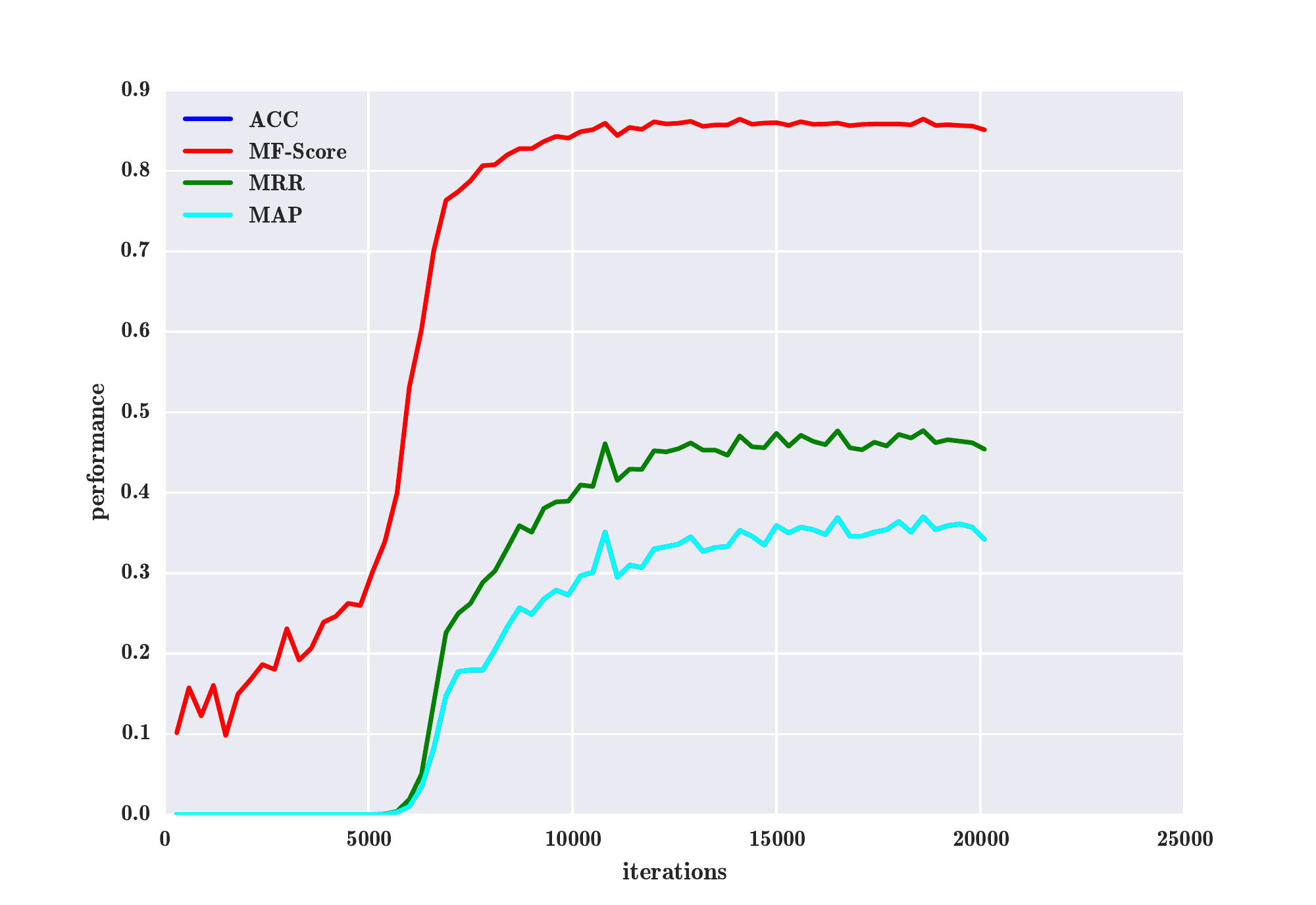}
        \caption{En-Hi}
        \label{fig:en-hi}
    \end{subfigure}
    \begin{subfigure}[b]{0.3\textwidth}
        \includegraphics[width=\textwidth]{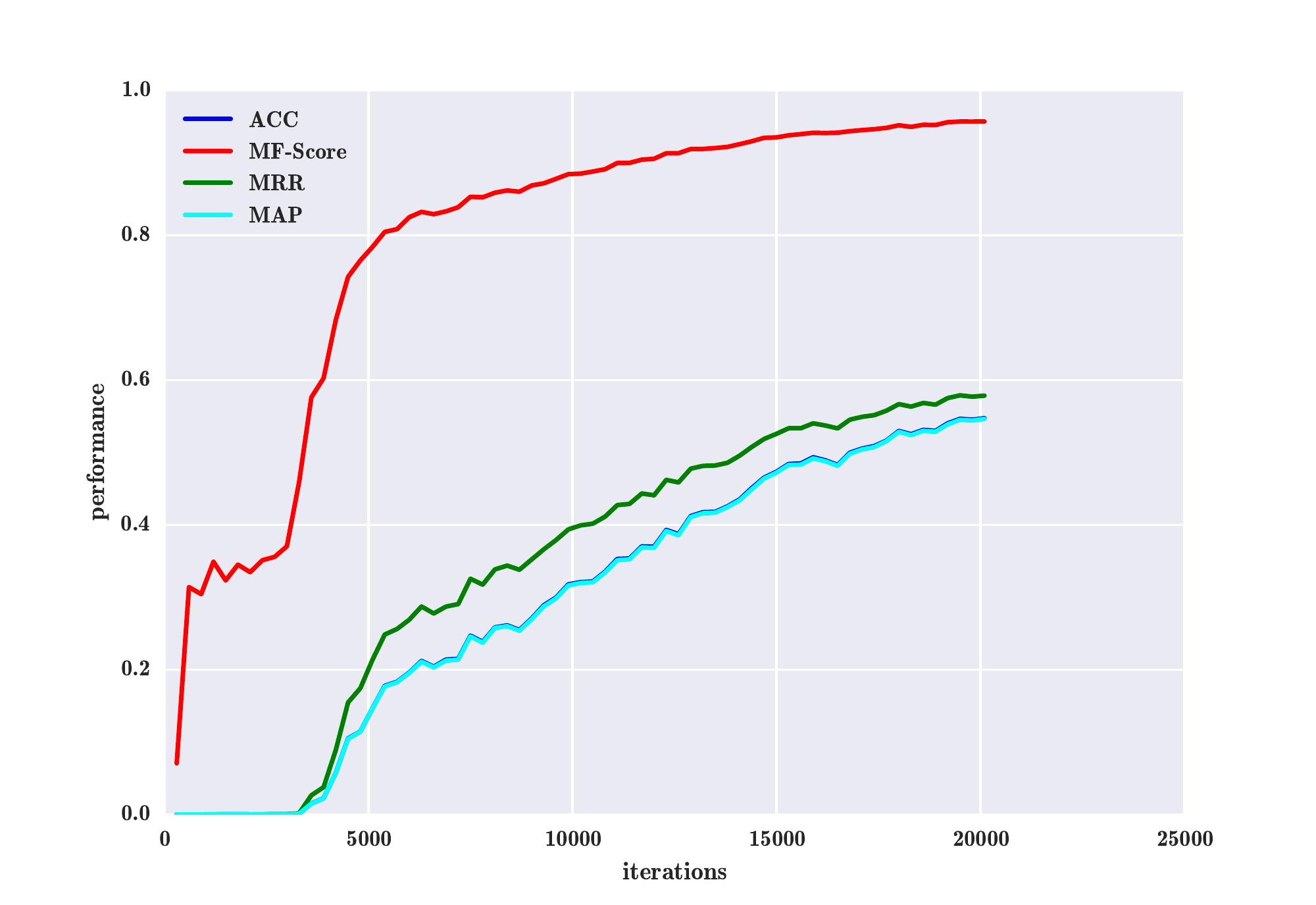}
        \caption{En-Ta}
        \label{fig:en-ta}
    \end{subfigure}
    \begin{subfigure}[b]{0.3\textwidth}
        \includegraphics[width=\textwidth]{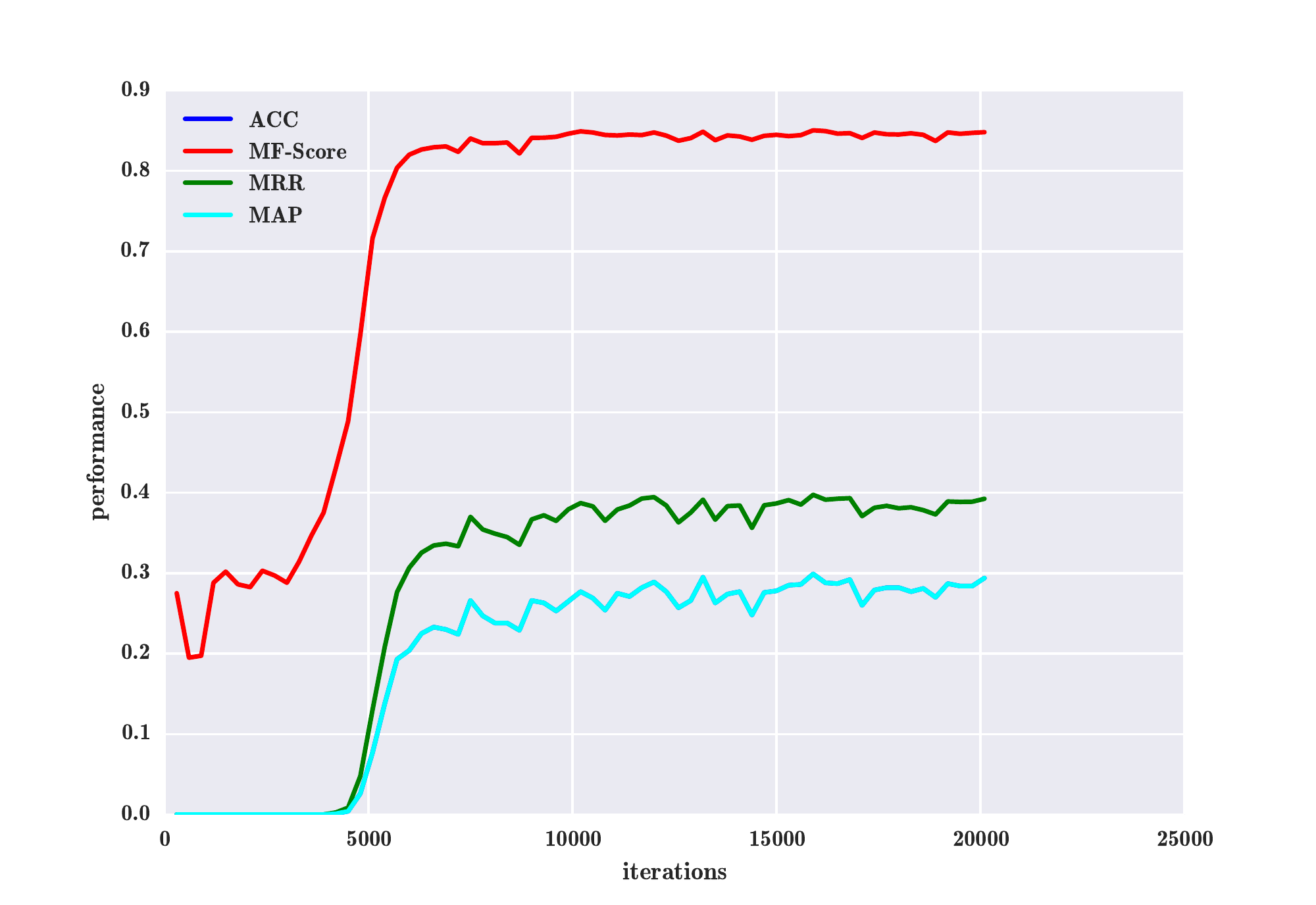}
        \caption{En-Ka}
        \label{fig:en-ka}
    \end{subfigure}
    \begin{subfigure}[b]{0.3\textwidth}
        \includegraphics[width=\textwidth]{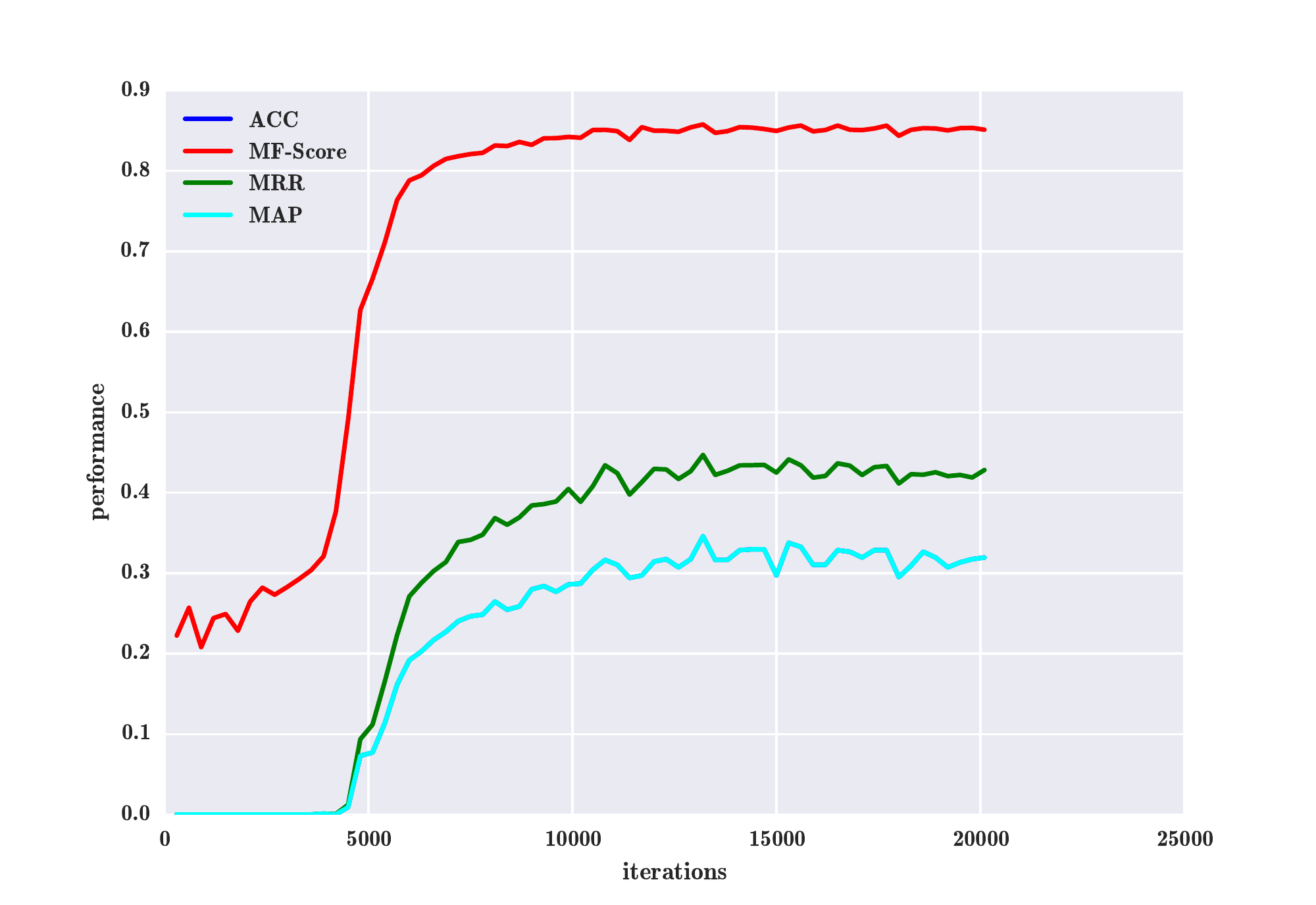}
        \caption{En-Ba}
        \label{fig:en-ba}
    \end{subfigure}
    \begin{subfigure}[b]{0.3\textwidth}
        \includegraphics[width=\textwidth]{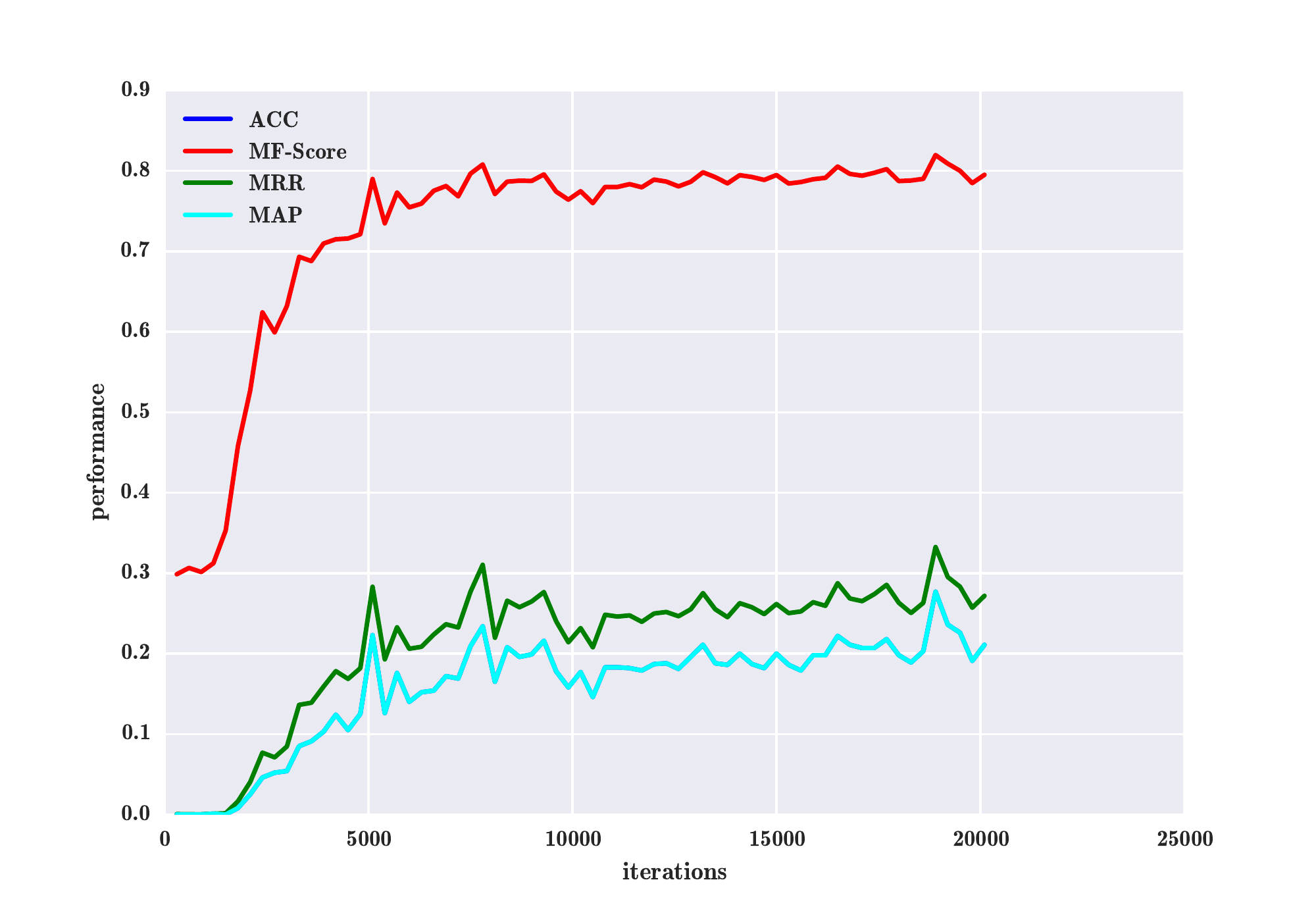}
        \caption{En-He}
        \label{fig:en-he}
    \end{subfigure}
    \caption{Learning curve of the proposed model on different datasets using the validation set. In most cases, the difference between 'ACC' and 'MAP' is negligible.}
    \label{fig:learningcurve}
\end{figure*}

Figure~\ref{fig:learningcurve} shows the learning curve of the proposed model on different datasets. It is clear that in most datasets, the trained model is capable of robust transliteration after a few number of iterations. As shown in Table~\ref{tbl:datasets}, each dataset has different number of training set and also different number of characters in the source and target language. For example, when transliterating from English to Chinese (TaskID=`En-Ch') and English to Hebrew, the target names contains $548$ and $37$ different tokens respectively. Since we leverage a same model for different datasets without tuning the model for each dataset, differences in the learning curves are expectable. For some datasets (such as `En-Ch'), it takes more time to fit the model to the training data while for some others (such as `En-He'), the model fit to the training data after a few iterations.

\section{Conclusion}
\label{sec:conclusion}
In this paper we proposed Neural Machine Transliteration based on successful studies in sequence to sequence learning~\cite{seq2seq14} and Neural Machine Translation~\cite{char_nmt15,char_nmt16,neural_translation15,nmt_properties14}. Neural Machine Transliteration typically consists of two components, the first of which encodes a source name sequence $\mathbf{x}$ and the second decodes to a target name sequence $\mathbf{y}$. Different parts of the proposed model jointly trained using stochastic gradient descent to minimize the log-likelihood. Experiments on different datasets using benchmark measures revealed that the proposed model is able to achieve significantly higher transliteration quality over traditional statistical models~\cite{smt10}. In this paper we did not concentrate on improving the model for achieving state-of-the-art results, so applying hyperparameter optimization~\cite{random_search12}, multi-task sequence to sequence learning~\cite{multi_task_seq2seq15} and multi-way transliteration~\cite{multi_nmt16,mtl_mt15} are quite promising for future works.

\section*{Acknowledgments}
The authors would like to thank the developers of Theano~\cite{theano16} and DL4MT~\footnote{\url{https://github.com/nyu-dl/dl4mt-tutorial}} projects. Also, the author would like to acknowledge the support of Bayan Inc. for research funding and computing support. The author also thank Yasser Souri for valuable comments.

\bibliography{references.bib}

\begin{thebibliography}{}

\bibitem[\protect\citename{Bahdanau \bgroup et al.\egroup
  }2015]{neural_translation15}
Dzmitry Bahdanau, Kyunghyun Cho, and Yoshua Bengio.
\newblock 2015.
\newblock Neural machine translation by jointly learning to align and
  translate.
\newblock In {\em International Conference on Learning Representations}, volume
  abs/1409.0473.

\bibitem[\protect\citename{Banchs \bgroup et al.\egroup }2015]{news15report}
Rafael~E. Banchs, Min Zhang, Xiangyu Duan, Haizhou Li, and A.~Kumaran.
\newblock 2015.
\newblock Report of news 2015 machine transliteration shared task.
\newblock In {\em Proceedings of the Fifth Named Entity Workshop}, pages
  10--23, Beijing, China, July. Association for Computational Linguistics.

\bibitem[\protect\citename{Bergstra and Bengio}2012]{random_search12}
James Bergstra and Yoshua Bengio.
\newblock 2012.
\newblock Random search for hyper-parameter optimization.
\newblock {\em J. Mach. Learn. Res.}, 13(1):281--305, February.

\bibitem[\protect\citename{Bojanowski \bgroup et al.\egroup
  }2015]{alt_char_rnn15}
Piotr Bojanowski, Armand Joulin, and Tomas Mikolov.
\newblock 2015.
\newblock Alternative structures for character-level rnns.
\newblock {\em arXiv preprint arXiv:1511.06303}.

\bibitem[\protect\citename{Cho \bgroup et al.\egroup }2014a]{nmt_properties14}
Kyunghyun Cho, Bart van Merrienboer, Dzmitry Bahdanau, and Yoshua Bengio.
\newblock 2014a.
\newblock On the properties of neural machine translation: Encoder--decoder
  approaches.
\newblock In {\em Proceedings of SSST-8, Eighth Workshop on Syntax, Semantics
  and Structure in Statistical Translation}, pages 103--111, Doha, Qatar,
  October. Association for Computational Linguistics.

\bibitem[\protect\citename{Cho \bgroup et al.\egroup
  }2014b]{learning_phrase_rnn14}
Kyunghyun Cho, Bart van Merrienboer, Caglar Gulcehre, Dzmitry Bahdanau, Fethi
  Bougares, Holger Schwenk, and Yoshua Bengio.
\newblock 2014b.
\newblock Learning phrase representations using rnn encoder--decoder for
  statistical machine translation.
\newblock In {\em Proceedings of the 2014 Conference on Empirical Methods in
  Natural Language Processing (EMNLP)}, pages 1724--1734, Doha, Qatar, October.
  Association for Computational Linguistics.

\bibitem[\protect\citename{Chung \bgroup et al.\egroup }2016]{char_decoder16}
Junyoung Chung, Kyunghyun Cho, and Yoshua Bengio.
\newblock 2016.
\newblock A character-level decoder without explicit segmentation for neural
  machine translation.
\newblock {\em CoRR}, abs/1603.06147.

\bibitem[\protect\citename{Costa{-}Juss{\`{a}} and Fonollosa}2016]{char_nmt16}
Marta~R. Costa{-}Juss{\`{a}} and Jos{\'{e}} A.~R. Fonollosa.
\newblock 2016.
\newblock Character-based neural machine translation.
\newblock {\em CoRR}, abs/1603.00810.

\bibitem[\protect\citename{Dong \bgroup et al.\egroup }2015]{mtl_mt15}
Daxiang Dong, Hua Wu, Wei He, Dianhai Yu, and Haifeng Wang.
\newblock 2015.
\newblock Multi-task learning for multiple language translation.
\newblock In {\em Proceedings of the 53rd Annual Meeting of the Association for
  Computational Linguistics and the 7th International Joint Conference on
  Natural Language Processing (Volume 1: Long Papers)}, pages 1723--1732,
  Beijing, China, July. Association for Computational Linguistics.

\bibitem[\protect\citename{Finch \bgroup et al.\egroup
  }2015]{nn_transliteration15}
Andrew Finch, Lemao Liu, Xiaolin Wang, and Eiichiro Sumita.
\newblock 2015.
\newblock Neural network transduction models in transliteration generation.
\newblock In {\em Proceedings of the Fifth Named Entity Workshop}, pages
  61--66, Beijing, China, July. Association for Computational Linguistics.

\bibitem[\protect\citename{{Firat} \bgroup et al.\egroup }2016]{multi_nmt16}
O.~{Firat}, K.~{Cho}, and Y.~{Bengio}.
\newblock 2016.
\newblock {Multi-Way, Multilingual Neural Machine Translation with a Shared
  Attention Mechanism}.
\newblock {\em ArXiv e-prints}, January.

\bibitem[\protect\citename{Hirschberg and Manning}2015]{nlp15}
Julia Hirschberg and Christopher~D. Manning.
\newblock 2015.
\newblock Advances in natural language processing.
\newblock {\em Science}, 349(6245):261--266.

\bibitem[\protect\citename{Jadidinejad and Mahmoudi}2010]{CLIR_meta10}
AmirHossein Jadidinejad and Fariborz Mahmoudi.
\newblock 2010.
\newblock Cross-language information retrieval using meta-language index
  construction and structural queries.
\newblock In Carol Peters, GiorgioMaria Di~Nunzio, Mikko Kurimo, Thomas Mandl,
  Djamel Mostefa, Anselmo Peñas, and Giovanna Roda, editors, {\em Multilingual
  Information Access Evaluation I. Text Retrieval Experiments}, volume 6241 of
  {\em Lecture Notes in Computer Science}, pages 70--77. Springer Berlin
  Heidelberg.

\bibitem[\protect\citename{Karimi \bgroup et al.\egroup
  }2011]{machine_transliteration_survey11}
Sarvnaz Karimi, Falk Scholer, and Andrew Turpin.
\newblock 2011.
\newblock Machine transliteration survey.
\newblock {\em ACM Comput. Surv.}, 43(3):17:1--17:46, April.

\bibitem[\protect\citename{Kingma and Ba}2014]{adam14}
Diederik~P. Kingma and Jimmy Ba.
\newblock 2014.
\newblock Adam: {A} method for stochastic optimization.
\newblock {\em CoRR}, abs/1412.6980.

\bibitem[\protect\citename{Koehn \bgroup et al.\egroup }2007]{moses07}
Philipp Koehn, Hieu Hoang, Alexandra Birch, Chris Callison-Burch, Marcello
  Federico, Nicola Bertoldi, Brooke Cowan, Wade Shen, Christine Moran, Richard
  Zens, Chris Dyer, Ond\v{r}ej Bojar, Alexandra Constantin, and Evan Herbst.
\newblock 2007.
\newblock Moses: Open source toolkit for statistical machine translation.
\newblock In {\em Proceedings of the 45th Annual Meeting of the ACL on
  Interactive Poster and Demonstration Sessions}, ACL '07, pages 177--180,
  Stroudsburg, PA, USA. Association for Computational Linguistics.

\bibitem[\protect\citename{Koehn}2010]{smt10}
Philipp Koehn.
\newblock 2010.
\newblock {\em Statistical Machine Translation}.
\newblock Cambridge University Press, New York, NY, USA, 1st edition.

\bibitem[\protect\citename{Ling \bgroup et al.\egroup }2015]{char_nmt15}
Wang Ling, Isabel Trancoso, Chris Dyer, and Alan~W. Black.
\newblock 2015.
\newblock Character-based neural machine translation.
\newblock {\em CoRR}, abs/1511.04586.

\bibitem[\protect\citename{Luong \bgroup et al.\egroup
  }2015]{multi_task_seq2seq15}
Minh{-}Thang Luong, Quoc~V. Le, Ilya Sutskever, Oriol Vinyals, and Lukasz
  Kaiser.
\newblock 2015.
\newblock Multi-task sequence to sequence learning.
\newblock {\em CoRR}, abs/1511.06114.

\bibitem[\protect\citename{Oh \bgroup et al.\egroup
  }2006]{transliteration_survey06}
Jong-Hoon Oh, Key-Sun Choi, and Hitoshi Isahara.
\newblock 2006.
\newblock A comparison of different machine transliteration models.
\newblock {\em J. Artif. Intell. Res. (JAIR)}, 27:119--151.

\bibitem[\protect\citename{Pascanu \bgroup et al.\egroup }2013]{deep_rnn13}
Razvan Pascanu, {\c{C}}aglar G{\"{u}}l{\c{c}}ehre, Kyunghyun Cho, and Yoshua
  Bengio.
\newblock 2013.
\newblock How to construct deep recurrent neural networks.
\newblock {\em CoRR}, abs/1312.6026.

\bibitem[\protect\citename{Sutskever \bgroup et al.\egroup }2014]{seq2seq14}
Ilya Sutskever, Oriol Vinyals, and Quoc~V. Le.
\newblock 2014.
\newblock Sequence to sequence learning with neural networks.
\newblock {\em CoRR}, abs/1409.3215.

\bibitem[\protect\citename{{Theano Development Team}}2016]{theano16}
{Theano Development Team}.
\newblock 2016.
\newblock {Theano: A {Python} framework for fast computation of mathematical
  expressions}.
\newblock {\em arXiv e-prints}, abs/1605.02688, May.

\bibitem[\protect\citename{Zhang \bgroup et al.\egroup }2015]{news15white}
Min Zhang, Haizhou Li, Rafael~E. Banchs, and A.~Kumaran.
\newblock 2015.
\newblock Whitepaper of news 2015 shared task on machine transliteration.
\newblock In {\em Proceedings of the Fifth Named Entity Workshop}, pages 1--9,
  Beijing, China, July. Association for Computational Linguistics.

\end{thebibliography}
\bibliographystyle{acl2016}

\end{document}